# Short Term Load Forecasting Models in Czech Republic Using Soft Computing Paradigms

Muhammad Riaz Khan and Ajith Abraham[*]

AMEC Technologies – Transtech Interactive Training Inc., 400-111 Dunsmuir Street, Vancouver, BC, V6B 5W3, Canada, E-mail: riaz.khan@amec.com

[*]Department of Computer Science, Oklahoma State University, Tulsa, OK 74106-0700, USA, Email: ajith.abraham@ieee.org

**Abstract**

*This paper presents a comparative study of six soft computing models namely multilayer perceptron networks, Elman recurrent neural network, radial basis function network, Hopfield model, fuzzy inference system and hybrid fuzzy neural network for the hourly electricity demand forecast of Czech Republic. The soft computing models were trained and tested using the actual hourly load data obtained from the Czech Electric Power Utility (CEZ) for the last seven years (January 1994 – December 2000). A comparison of the proposed techniques is presented for predicting 48 hourly (2 day ahead) demands for electricity. Simulation results indicate that hybrid fuzzy neural network and radial basis function networks are the best candidates for the analysis and forecasting of electricity demand.*

## 1. Introduction

Load forecasting is an essential element of power system operation and planning involving prognosis of the future level of demand to serve as the basis for supply-side and demand-side planning [2] [10] [14]. This includes planning for transmission and distribution facilities as well as new generation plants. Load forecasts are prepared for different time frames and levels of detail. An overall generation plan requires a system level forecast of total generation requirements and peak demand. Transmission and distribution planning, on the other hand, requires far more level and geographic details to assess the location, timing and loading of individual lines, substation and transformation facilities [1].

Statistical techniques like auto-regression and time-series methods being predominantly conventional have shown reasonably good results in the past. These conventional methods have the inherent inaccuracy of load prediction and numerical instability. Further, the non-stationarity of the load prediction process, coupled with complex relationship between weather variables and the electric load render such conventional techniques ineffective as these methods assume simple linear relationships during the prediction process.

Artificial Neural Networks (ANN) have the ability to learn and construct a complex nonlinear mapping through a set of input/output examples. ANN consist of a large number of parallel-processing units, which can be implemented using software or general-purpose neural network hardware. Fuzzy Systems (FS) exhibit complementary characteristics, offering a very powerful framework for approximate reasoning as it attempts to model the human reasoning process at a cognitive level. FS acquires knowledge from domain experts and this is encoded within the algorithm in terms of the set of *If-Then* rules. Fuzzy systems employ this rule based approach and interpolative reasoning to respond to new inputs.

For developing the forecasting models, we used the actual hourly electrical load data provided by the Czech Electric Power Utility (CEZ) for the years 1994 through 1999. The weather parameters temperature, humidity, wind speed and wind chill affect the forecasting accuracy during summer and winter. The input parameters considered for training the models were maximum, minimum and average temperature, humidity, wind speed and wind-chill, respectively. To ascertain the forecasting accuracy, the developed models were tested/evaluated on the data for the year 2000.

The paper is organized as follows. In section 2, we give a brief overview of load pattern in the Czech Republic and the factors affecting the load demand. Section 3 discusses the modeling of input data to train the different forecast models. In section 4, a short theoretical background of all forecasting models is presented followed by test results and discussion in section 5. Conclusions are drawn in section 6.

## 2. Demand Patterns in Czech Republic

A broad spectrum of factors affects the system's load level such as trend effects, cyclic-time effects, special effects, weather effects, random effects like human activities, load management, pricing strategy, electricity tariff structures, visibility, illumination level and thunderstorms. In addition, total system load is subjected to random disturbances caused by sudden increase of large loads or outages. Thus the load profile is dynamic in nature with temporal, seasonal and annual variations.

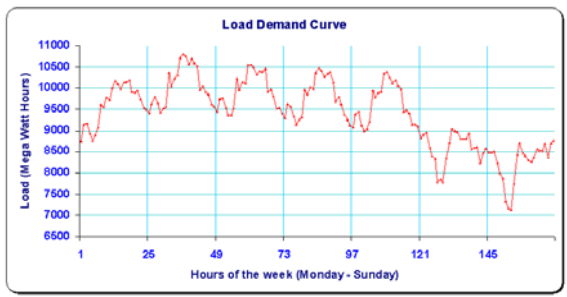

**Figure 1:** Typical weekly load curve in the Czech Republic.

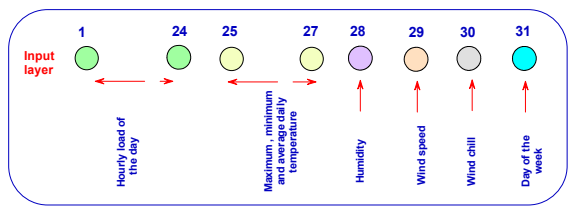

**Figure 2:** Input data scheme for 24 hour ahead load forecasting.

The electrical peak consumption of the Czech Republic is about 1 GW during a working day in winter. This demand is highly volatile on a day-to-day basis and is being significantly affected by weather conditions. The weekly pattern is comprised of the daily shapes (Monday through Sunday), reflecting the main working activities. Figure 1 shows the shapes of the typical electrical load curve during a week. Generally the load pattern on normal weekdays, when work is already in full swing, remains almost constant with small random variations from varying industrial activities, weather conditions, etc. The load values for normal days are functions of the short-term historical data and forecast values of weather parameters (e.g. high and low temperatures). The load on Mondays and Fridays is different from that on other weekdays due to pickup loads on Monday mornings when all business and industries just start work, and evening loads on Fridays, because of its proximity to the weekend. The load pattern on Saturdays is different from rest of the weekdays. The peak load also takes a dip on Saturdays, which is a rest day for most of the people. The shape of the load curve on Sundays is similar to that on holidays. The peak load decreases considerably before and after major public holidays.

## 3. Input Variable Selection and Modeling

The most important work in building our soft computing based Short Term Load Forecasting (STLF) models is the selection of the input variables. Actually, there is no guaranteed rule that one could follow in this process. It mainly depends on experience and is carried out almost entirely by trail and error. However, some statistical analysis can be very helpful in determining the variables, which have significant influence on the system load. Normally more input neurons make the performance of the neural network worse in many circumstances. We had to use extra input neurons to represent the necessary weather parameters, which have strong correlation with the electric load.

For two-days ahead load forecasting, we used 62 inputs nodes: the first 48 nodes represent the past 48 hour loads, nodes # 49-54 are used for maximum, minimum and average temperature for the past two days temperature, the remaining 8 nodes represent humidity, wind-speed, wind-chill and days of the week. Input data scheme for the hourly prediction (24 hour ahead) is depicted in Figure 2. The hidden layer of Multi Layer Perceptron (MLP), Elman Recurrent Neural Network (ERNN) and Radial Basis Function Network (RBFN) consists of 24, 60 and 298 hidden neurons, respectively as shown in Table 1. This number was determined from studying the network behavior during the training process taking into consideration some factors like convergence rate, error criteria etc. The output layer consists of 48 neurons each representing the predicted hourly load of two days.

## 4. Soft Computing Models

Soft Computing (SC) introduced by Lotfi Zadeh [13] is an innovative approach to construct computationally intelligent hybrid systems consisting of Artificial Neural Network (ANN), Fuzzy Logic (FL), approximate reasoning and derivative free optimization methods such as Evolutionary Algorithms (EAs). In this section, a brief theoretical background of the different soft computing models considered is given.

### 4.1. Multilayer Perceptron Network

We used a fully connected feedforward type neural network (Figure 3) consisting of one hidden layer.

Backpropagation algorithm was utilized for training the MLP network. The training error level was set to $10^{-4}$. The optimal number of hidden neurons was obtained experimentally by changing the network design and running the training process several times until a good performance was obtained [6].

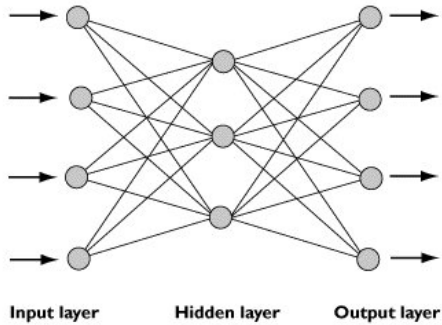

**Figure 3.** Feedforward neural network.

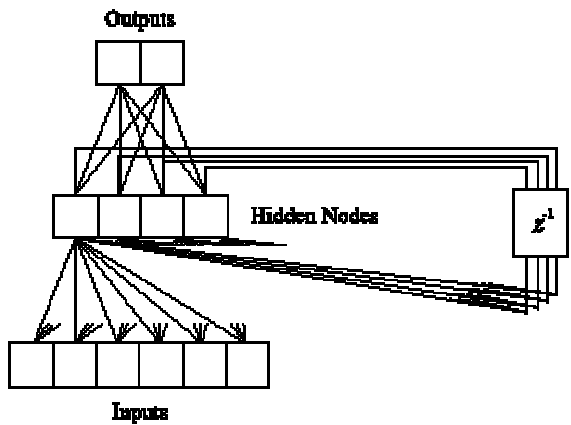

**Figure 4.** Elman recurrent neural network.

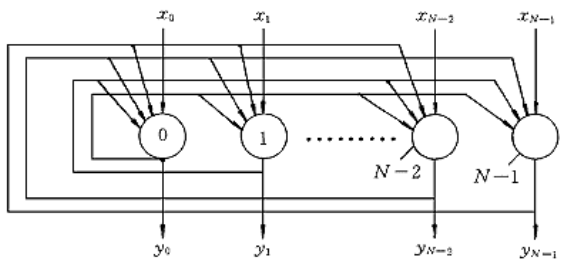

**Figure 5.** Hopfield neural network.

### 4.2. Elman Recurrent Neural Network

Recurrent neural networks, being member of a class of neural network models exhibiting dynamic behavior, are often used to represent dynamical systems [12]. Due to the nonlinear nature of these models, the behavior of the load prediction system can be captured in a compact, robust and more natural representation. We used the Elman network (Figure 4) also known as partial recurrent network or simple recurrent network with one hidden layer. In this network, the outputs of the hidden layer are allowed to feedback onto itself through a buffer or context layer. This feedback allows Elman networks to learn to recognize and generate temporal patterns, as well as spatial patterns. Every hidden neuron is connected to only one neuron of the context layer through a constant weight of value one. Hence, the context layer constitutes a kind of copy or memory of the state of the hidden layer, one instant before. The number of context neurons is consequently the same as the number of hidden neurons. Every neuron in the hidden layer receives as input, in addition to the external inputs of the network, the outputs of the context layer neurons. Inputs, output and context neurons have linear activation functions while hidden neurons have sigmoidal activation function.

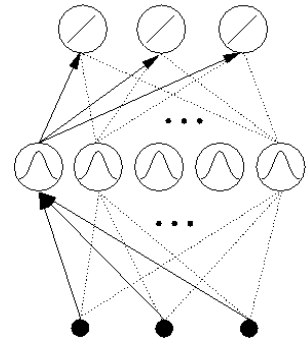

**Figure 6.** Radial Basis Function Network

### 4.3. Radial Basis Function Network

RBFNs exhibit a good approximation and learning ability and are easier to train and generally converge very fast. The RBFN is a 3-layered feedforward network (Figure 6) comprising of input, hidden/memory, and output neurons respectively. It uses a linear transfer function for the output units and Gaussian function (radial basis function) for the hidden units [5].

### 4.4. Hopfield Model

This network is a single layer network (Figure 5) with symmetric weight matrices in which the diagonal elements are all zero. The diagonal elements need not be zero, but we assume that is the case since the performance is improved when taken to be zero. Thus, for a Hopfield network with weight matrix $W$, $w_{ij} = w_{ji}$ and $w_{ii} = 0$ for all $i, j = 1, 2, ..., n$. Inputs are applied simultaneously to all neurons, which then output to each other and the process continues until a stable state is reached, which represents the network output [4].

**Table 1.** Comparison of training parameters in RBFN, MLP and ERNN networks.

| Soft computing model | Number of hidden neurons | Training Time (minutes) | Activation function used in hidden layer | Activation function used in output layer |
|---|---|---|---|---|
| RBFN | 298 | 10 | Gaussian function | Linear |
| MLP | 24 | 240 | Tan-sigmoidal | Linear |
| ERNN | 60 | 360 | Log-sigmoidal | Linear |

**Table 2:** MAPE and MAP for working days of a week using various MFs.

| Working Days | Membership Functions | | | | | |
|---|---|---|---|---|---|---|
| | Triangular | | Gaussian Curve | | Trapezoidal | |
| | MAPE (%) | MAP (%) | MAPE (%) | MAP (%) | MAPE (%) | MAP (%) |
| **Monday** | 2.60 | 6.30 | 2.72 | 7.85 | 2.84 | 5.78 |
| **Tuesday** | 1.40 | 4.95 | 1.44 | 4.69 | 2.14 | 6.64 |
| **Wednesday** | 0.98 | 3.67 | 1.62 | 4.98 | 1.87 | 4.65 |
| **Thursday** | 0.89 | 4.53 | 1.81 | 5.30 | 2.99 | 5.25 |
| **Friday** | 1.20 | 3.54 | 1.48 | 4.39 | 2.29 | 4.99 |

**Table 3:** MAPE and MAP for one weekend using various membership functions.

| Weekend Days | Membership Functions | | | | | |
|---|---|---|---|---|---|---|
| | Triangular | | Gaussian Curve | | Trapezoidal | |
| | MAPE (%) | MAP (%) | MAPE (%) | MAP (%) | MAPE (%) | MAP (%) |
| **Saturday** | 3.16 | 8.05 | 4.34 | 7.18 | 5.48 | 8.66 |
| **Sunday** | 3.18 | 7.16 | 4.19 | 8.62 | 5.64 | 9.84 |

The feedback loops involve the use of particular branches composed of unit-delay elements (denoted by $z^{-1}$), which result in a nonlinear dynamical behavior by virtue of the nonlinear nature of the neurons.

### 4.5. Fuzzy Inference System

If we use appropriate membership function definitions and a well-defined rule base, we can achieve good prediction accuracy [3] [8] [9]. Fuzzy systems are stable, easily tunable and could be validated conventionally. In our experiments, two years of historical load and weather data were used, one year (1999) for designing the fuzzy rule base design and the following year (2000) for testing the model performance. We used a Mamdani fuzzy inference system [7] for predicting the 24-hour ahead (weekdays and weekends) load demand. To ensure prediction accuracy, the number of fuzzy membership functions and shape of the fuzzy membership functions were changed and new fuzzy rule base was obtained. The iterative process of designing the rule base, choosing a defuzzification algorithm, and testing the system performance was repeated several times with a different number of fuzzy membership functions and different shapes of fuzzy memberships.

The fuzzy rule base that provided the minimum error measure for the test set was selected for real-time forecasting. We used various Membership Functions (MF) such as triangular, trapezoidal, Gaussian-curve and bell-shaped. Using different MF, the mean absolute percentage error (MAPE) and maximum absolute percentage error (MAP) for working days of the week and weekend are computed and are depicted in Tables 3 and 4 respectively. Empirical values from Tables 2 and 3 depicts that the selection of different MFs e.g., triangular, Gaussian, trapezoidal etc. significantly affect the prediction performance.

## 4.6. Hybrid Fuzzy Neural Network

A hybrid fuzzy-neural network approach, which combines the important features of ANN and fuzzy logic, is also proposed in this paper. This architecture is suggested for realizing cascadable fuzzy inference system and neural network modules, which are used as building blocks for constructing a load forecasting system. Expert knowledge represented by fuzzy rules is used for preprocessing input data fed to an ANN. In order to train the ANN for 48 hours ahead load forecasting, fuzzy *if-then* rules are used, in addition to historical load and weather data that are usually employed in conventional supervised learning methods as shown in Figure 7. The fuzzy membership values of load and other weather variables are the inputs to the ANN and the output comprises the membership value of the predicted load. To deal with the linguistic values such as high, low, and medium, architecture of ANN that can handle fuzzy input vectors is propounded.

The FNN is trained on real data provided by the CEZ and evaluated for forecasting 48 hours load profiles. We used a Takagi-Sugeno fuzzy model [11], as it is capable of representing the dynamics of a complex system using fewer fuzzy rules.

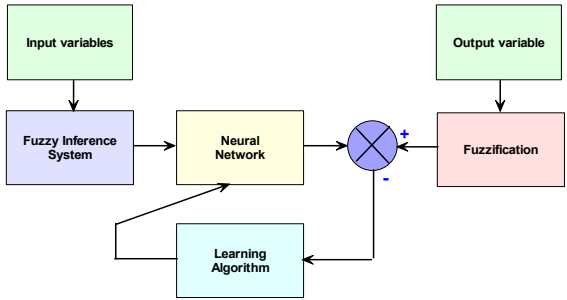

**Figure 7** Structure of the FNN architecture during training.

Membership function defines a fuzzy set by mapping crisp inputs from its domain to degrees of membership. Each input variable is converted into a fuzzy membership function in the range [0 – 1] that corresponds to the degree to which the input belongs to a linguistic class. In our research, we used the Gaussian MFs for both load and temperature inputs to fuzzify the linguistic variables. Numerical values of the membership function $\mu_p$ translated from its linguistic representation for each variable $P$ are calculated as:

$$\mu_p = \begin{cases} 1 & if \ P = B \\ [P - A]/[B - A] & if \ B > P > A \\ [C - P]/[C - B] & if \ B < P < C \\ 0 & if \ P \geq C \ or \ P \leq A \end{cases}$$

The input load is sorted into 7 categories and labeled as extremely low (ExL), very low (VL), low (L), normal (N), high (H), very high (VH), and extremely high (ExH). The input temperature is sorted into 8 categories and labeled as extremely cold (ExC), very cold (VC), cold (C), normal (N), warm (W), hot (H), very hot (VH), and extremely hot (ExH). The humidity is sorted into seven categories and labeled as extremely low (ExL), very low (VL), low (L), medium (M), high (H), very high (VH) and extremely high (ExH). The wind speed is labeled as zero (Z), positive very small (PVS), positive small (PS), medium (M), positive medium (PM), big (B) and positive big (PB). Similarly, wind chill is labeled as zero (Z), very very low (VVL), very low (VL), low (L), high (H), very high (VH) and extremely high (ExH) as shown in Figure 8.

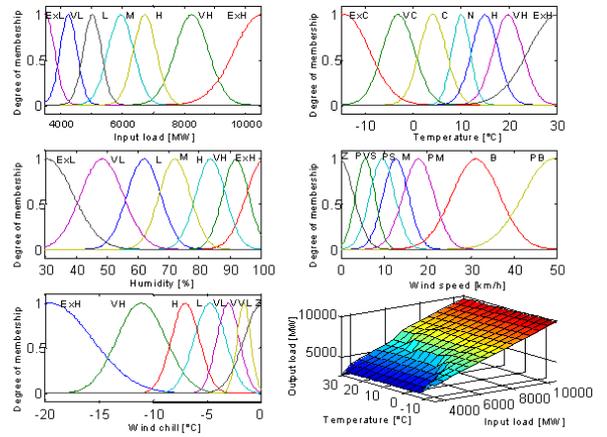

**Figure 8.** Input parameters using Gaussian-curve membership function.

A Takagi-Sugeno type fuzzy inference system is generally represented by [11]

*If $P_1$ is $\mu_1$ and $P_2$ is $\mu_2$ and ... $P_n$ is $\mu_n$, then $f_1 = Q_0 + P_1Q_1 + P_2Q_2 + ... + P_nQ_n$*

The truth-value of each rule is obtained by

$$w_1 = (\mu_1 \wedge \mu_2 \wedge ... \wedge \mu_n)$$

where $P_i$ are the premise variables, $\mu_i$ the membership functions and $f_i$ the consequence of the *i*th rule whose value is inferred based on parameters $Q_i$ when $P_i$ satisfies the premise. Some of the terms may or may not appear in each rule. The final value of *f* is obtained as the weighted combination of results from all such rules

$$f = (\sum w_i f_i / \sum w_i) \quad \text{for all } i = 1, ..., n$$

The ANN is allowed to train until it maps the input-output relationship with the desired accuracy. At the final stage, the output from the neural network is a fuzzified set of data, which indicate the degree of membership of the outcome in the range $[dP = P_{max} - P_{min}]$. This result is defuzzified by converting it in the pre-specified range to obtain the hourly load values (in MW) for every hour of the day.

$$P = P_{min} + P_n * dP$$

where $P_n$ is the predicted output load from the ANN and $P$ is the corresponding load value in MW.

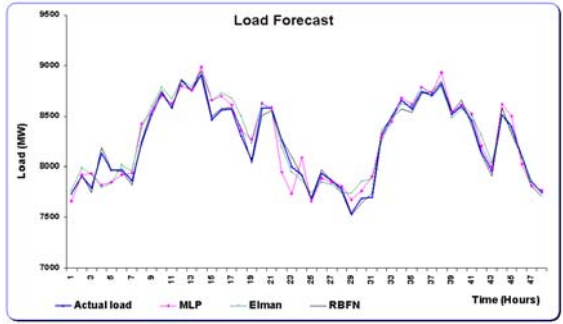

**Figure 9.** Comparison of 48 hrs ahead forecast for working days (Thursday and Friday) using MLP, Elman and RBFN networks.

## 4.7 Test Results and Discussions

The assessment of the prediction performance of the different soft computing models were done by quantifying the prediction obtained on an independent data set. The maximum absolute percentage error (*MAP*) and mean absolute percentage error (*MAPE*) were used to study the performance of the trained forecasting models for the testing year 2000.

*MAP* is defined as follows:

$$MAP = max\left(\frac{|P_{actual,\,i} - P_{predicted,\,i}|}{P_{predicted,\,i}} \times 100\right)$$

where $P_{actual,\,i}$ is the actual load on day $i$ and $P_{actual,\,i}$ is the forecast value of the load on that day.

Similarly *MAPE* is given as

$$MAPE = \frac{1}{N} \sum_{i=1}^{N} \left[\frac{P_{actual,\,i} - P_{predicted,\,i}}{P_{actual,\,i}}\right] \times 100$$

where $N$ represents the total number of hours.

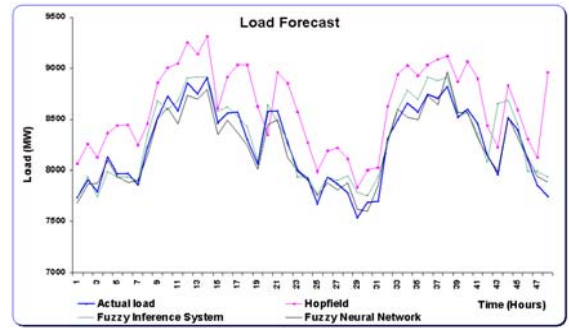

**Figure 10.** Comparison of 48 hrs ahead forecast for working days (Thursday and Friday) using Hopfield model, fuzzy inference system and FNN.

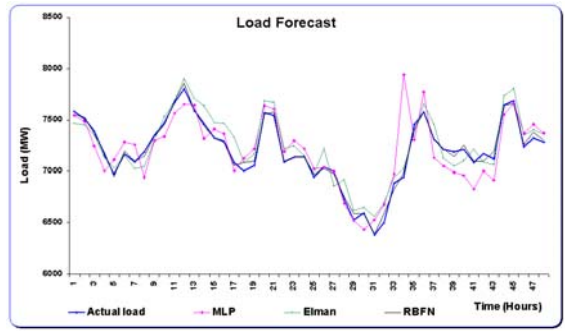

**Figure 11.** Comparison of 48 hrs ahead forecast for weekend days (Saturday and Sunday) using MLP, Elman and RBFN networks.

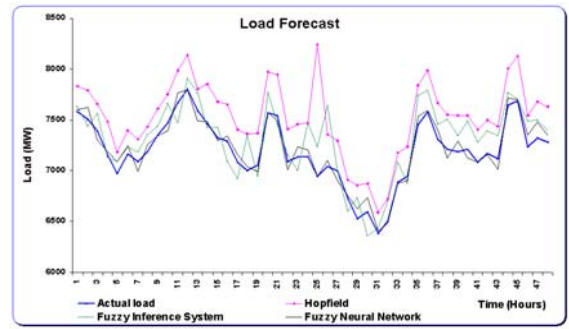

**Figure 12.** Comparison of 48 hrs ahead forecast for weekend days (Saturday and Sunday) using Hopfield model, Fuzzy logic and FNN.

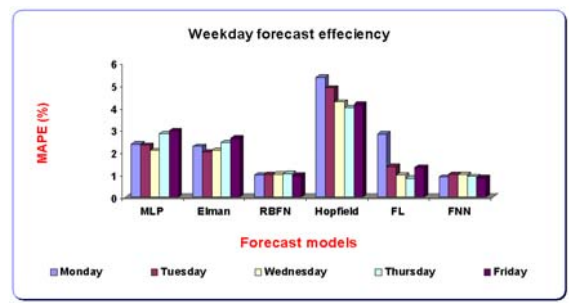

**Figure 13.** Mean absolute percentage error (MAPE) computed for working days of one week.

**Table 4.** MAPE of 24 hour forecast during weekdays.

| Model | Monday | Tuesday | Wednesday | Thursday | Friday |
|---|---|---|---|---|---|
| **MLP** | 2.360 | 2.310 | 2.070 | 2.830 | 2.960 |
| **ERNN** | 2.260 | 2.010 | 2.090 | 2.430 | 2.640 |
| **RBFN** | 0.980 | 1.000 | 1.020 | 1.060 | 0.960 |
| **Hopfield** | 5.340 | 4.870 | 4.230 | 3.990 | 4.120 |
| **FL** | 2.810 | 1.360 | 0.980 | 0.840 | 1.320 |
| **FNN** | 0.890 | 1.010 | 1.000 | 0.920 | 0.870 |

**Table 5.** MAPE of 24 hour forecast during weekends.

| Model | Saturday | Sunday |
|---|---|---|
| **MLP** | 2.48 | 2.5 |
| **ERNN** | 2.75 | 2.86 |
| **RBFN** | 1.26 | 1.38 |
| **Hopfield** | 5.98 | 6.10 |
| **FL** | 2.99 | 2.74 |
| **FNN** | 2.01 | 1.99 |

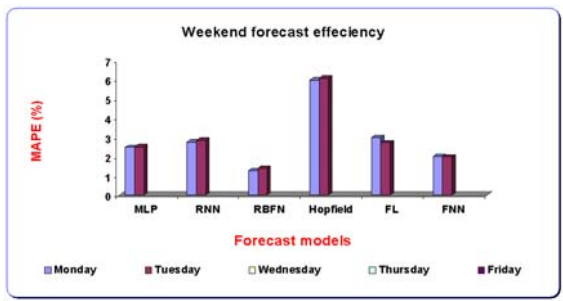

**Figure 14.** Mean absolute percentage error (MAPE) computed for weekend days of one week.

We used a Pentium, 300 MHz platform for simulating the prediction models using MATLAB version 5.3. In order to show the efficacy of the forecasting models, the hourly forecast results for both working days and holidays are shown in Figures 9, 10, 11 and 12 respectively. Comparison of computed MAPE for all forecasting models for working days and weekend of a week are also depicted in Figures 13 and 14, respectively. The empirical values of the test results are depicted in Tables 4 and 5 respectively.

This paper discusses only hourly load forecasting with lead-time of 48 hours. Two separate models were developed for weekdays and weekend load forecasting, although utility companies have little interest in weekend load forecasting since the load demand of weekend is much lesser than that of weekdays.

## 5. Conclusions

A comparative study of soft computing models for load forecasting shows that FNN and RBFN are more accurate and effective as compared to MLP, ERNN, Hopfield model and a simple fuzzy inference system. The error associated with each method depends on several factors such as homogeneity in data, choice of model, network parameters, and finally the type of solution.

ANNs have gained great popularity in time-series prediction because of their simplicity and robustness. The learning method is normally based on the gradient descent method – backpropagation algorithm. Backpropagation algorithm has major drawbacks: the learning process is time-consuming and there is no exact rule for setting the number of hidden neurons to avoid overfitting or underfitting, and hopefully, making the learning phase convergent. In order to eliminate such problems, the RBFN has been applied. The results obtained clearly demonstrate that RBFN are much faster and more reliable for short term load forecasting.

The ANN based approach is not the only way to predict the short-term load demand, nor it is necessarily the best way for all purposes. A forecasting technique based on FL approach has also been presented as an alternative technique. The flexibility of the FL approach, offering a logical set of *if-then* rules, which could be easily understood by an operator, might be a very good solution for easy practical implementation and usage of STLF models.

The hybrid FNN approach was finally used to forecast loads with greater accuracy than the conventional approaches when used on a stand-alone mode.

The knowledge base was easily developed and modified to reflect changes in weather-load relationship during different seasons. FNN training time was much faster than ANN and also effectively incorporated linguistic *if-then* expert rules. FNN provided a general method for combining available numerical information and human linguistic information in a common framework. Significant accuracy was achieved due its efficient adaptive tracking capability that results in the development of a robust and accurate forecasting technique. It may be concluded that while each method has its own advantages and disadvantages, each method in its own merit is more accurate than conventional statistical techniques and hence worthy of consideration to the application of load forecasting.

## Acknowledgements

The authors would like to thank the Czech Power Company (CEZ), Prague, Czech Republic for providing the electrical load and temperature data used in this work. This research was possible with support of Vyzkumny Zámer CEZ: J22/98:262200010, Czech Republic.